\RequirePackage{fix-cm}
\documentclass[smallextended]{svjour3}       
\smartqed  
\usepackage{hyperref}
\usepackage{color}

\usepackage{graphicx}
\usepackage{mathptmx}      
\usepackage{amsmath,amssymb}
\usepackage{subfig}
\usepackage{natbib}
\usepackage{url}
\usepackage[ruled,lined,titlenumbered,linesnumbered]{algorithm2e}

\newcommand{\entrances}{\ensuremath{\varepsilon}}
\newcommand{\objectives}{\ensuremath{\phi}}
\newcommand{\detectors}{\ensuremath{\delta}}

\newcommand{\radius}{\ensuremath{\tau}}
\newcommand{\neutral}{\ensuremath{\theta}}

\newcommand{\Greedy}{Greedy}
\newcommand{\HC}{Hill Climbing}
\newcommand{\EA}{Evolutionary Algorithm}

\newcommand{\refRange}[3]{{#1}~\ref{#2}--\ref{#3}}
\newcommand{\refAndPage}[2]{{#2}~\ref{#1}}
\newcommand{\refTable}[1]{\refAndPage{#1}{Table}}
\newcommand{\refFig}[1]{\refAndPage{#1}{Figure}}
\newcommand{\refAlg}[1]{\refAndPage{#1}{Algorithm}}
\newcommand{\refFigs}[2]{\refRange{Figures}{#1}{#2}}

\newcommand{\refEq}[1]{\refAndPage{#1}{Equation}}

\newcommand{\refSect}[1]{\refAndPage{#1}{Section}}

\newcommand{\harbour}{\textsf{harbour}}
\newcommand{\newtown}{\textsf{newtown}}
\newcommand{\oldtown}{\textsf{oldtown}}

\DeclareMathOperator*{\argmin}{arg\,min}
\DeclareMathOperator*{\argmax}{arg\,max}

\newcommand{\undermax}[1]{\underset{#1}{\max}}
\newcommand{\underargmax}[1]{\underset{#1}{\argmax}}
\newcommand{\underargmin}[1]{\underset{#1}{\argmin}}
\journalname{Natural Computing}

\newcommand{\fnctn}[1]{\textsc{#1}}
\newcommand{\kw}[1]{\normalfont \textbf{#1}}

\begin{document}

\title{New Perspectives on the Optimal Placement of Detectors for Suicide Bombers using Metaheuristics\thanks{This work is
  supported by the Spanish Ministerio de Econom\'{\i}a and European FEDER under 
  Projects EphemeCH (TIN2014-56494-C4-1-P) and 
  DeepBIO (TIN2017-85727-C4-1-P).}
\thanks{This document is a preprint of Cotta, C., Gallardo, J.E. New perspectives on the optimal placement of detectors for suicide bombers using metaheuristics. Nat Comput 18, 249–263 (2019). \url{https://doi.org/10.1007/s11047-018-9710-1}}
}

\titlerunning{New Perspectives on the OPSBD using Metaheuristics}        

\author{Carlos Cotta \and Jos\'e E. Gallardo}


\institute{C. Cotta \and J.E. Gallardo  \at
              ETSI Inform\'atica, Universidad de M\'alaga, Campus de Teatinos, 29071 M\'alaga (Spain) \\
              Tel.:  +34 952137158 / +34 952132795 \\
              Fax: +34 952131397\\
              \email{\{ccottap,pepeg\}@lcc.uma.es}           
}

\date{Received: date / Accepted: date}

\maketitle

\begin{abstract}
We consider an operational model of suicide bombing attacks --an increasingly prevalent form
of terrorism-- against specific targets, and the use of protective countermeasures based on the deployment 
of detectors over the area under threat. These detectors have to be carefully located in order to 
minimize the expected number of casualties or the economic damage suffered, resulting in a
hard optimization problem for which different metaheuristics have been proposed. Rather than assuming
random decisions by the attacker, the problem
is approached by considering different models of the latter, whereby he takes informed
decisions on which objective must be targeted and through which path it has to be reached based
on knowledge on the importance or value of the objectives or on the defensive strategy of the defender
(a scenario that can be regarded as an adversarial game). We consider four different algorithms,
namely
a greedy heuristic, a hill climber, tabu search and an evolutionary algorithm, and study their
performance on a broad collection of problem instances trying to resemble different realistic settings
such as a coastal area, a modern urban area, and the historic core of an old town. It is shown that 
the adversarial scenario is harder for all techniques, and that the evolutionary algorithm seems to
adapt better to the complexity of the resulting search landscape.


\end{abstract}

\section{Introduction}
\label{sec:introduction}

The threat of international terrorism has been present for many decades now, but ever since
the collapse of the Eastern bloc in the late 1980s and the subsequent end of the Cold War it has emerged
as the major factor of instability and social disruption at a global scale, be it directly as
a consequence of terrorist actions or indirectly as a result of the countermeasures required
to prevent the former and the impact of these in our daily lives (not to mention the
military operations targeting terrorist groups and regimes supporting them, which in
themselves contribute both to local turmoil and social unrest). Indeed, the new 
international environment arising from the end of the Atlantic/Eastern
bloc dialectic (e.g., easier and increased communications and travel, larger citizen mobility,
etc.) and the technical advances that took place in the end of the 20th century were readily
exploited by terror groups to bring their criminal actions to a new level. As a response, 
democratic nations have typically engaged in a counterterrorist policy which can be described 
within the framework of the 4D
strategy \citep{CIA03national}: \emph{Defeat} (proactively attack the structure of terrorist organizations), \emph{Deny} 
(preclude
the access to any kind of support --political, financial, etc.-- and resources for undergoing
terrorist attacks), \emph{Diminish} (minimize the underlying conditions that serve as a breeding
ground for the emergence of terrorists groups) and \emph{Defend} (protect the society, 
the citizens and their interests).

We are particularly concerned here with the last of the four Ds, namely
the defensive component. This constitutes a challenge in light of the
non-conventional means used by terrorists to conduct their infamous 
actions. Most conspicuously, suicide bombings have established themselves
as a particularly important menace \citep{cpost16database}, given their lethality (they cause
about four time more casualties than other kinds of terrorist attacks
\citep{rand09database,edwards1640years}), simplicity (their nature makes escaping logistics
and equipment for remote operation unneeded), and effectiveness in
instilling fear and cause social disruption \citep{hoffman03logic}.
Defense from this kind of attacks (and from any other of terrorist
attack for that matter) is a highly cross-disciplinary endeavor that can be
approached from many points of views. One of them is undoubtedly
the understanding of these actions, not just of the means but also of
the motivations. While the later can be contentious in nature, it can
be nevertheless approached at a meta-level, focusing not just on the
precise motivations but on the framework in which these motivations
substantiate into actions. In this sense, there are two main schools
of thought: the psychological-sociological and the political-rational approach
\citep{ganor11trends}. While the former focus on social group dynamics and the psychological
profile of individuals, the latter views terrorism as a rational method
of operation for pursuing concrete political goals. Without denying the
interest of the former (which can be useful for, e.g., detecting radicalization
of particular individuals or understanding their activity patterns --
see \citep{lara17measuring,leistedt16radicalization,tutun17framework}),
the latter can be particularly interesting from a tactical point of view, that is,
when it comes to implementing measures to make attacks unsuccessful. Indeed,
many details of a terrorist attack can be dictated by rational calculations
and cost-benefit analysis, much like it would be done by a military staff
or a corporate board \citep{blomberg04economic}.

In line with the previous line of reasoning, both the actions of the terrorists
and the countermeasures taken to neutralize the former can be regarded as
aimed to optimize (in different senses, obviously) some utility function, such as
the number of civilian casualties or the economic damage caused in the
infrastructures. This
approach has been precisely considered in some works in the literature. Thus,
\citet{Xiaofeng2007} considered --following 
the seminal work by \citet{kaplan05operational}-- the case of an area under threat of a suicide
bombing attack and the deployment of some kind of sensors (tailored to detect traces of
explosives in their proximity \citep{gares16review}) in order to protect specific known targets. To this
end, they proposed a branch-and-bound (BnB) algorithm and a greedy constructive
heuristic to determine the location of these sensors so as to
minimize the casualties. This same approach was also applied 
by \citet{Yan201671} to a scenario involving maritime targets under 
threat of small vessels. Following these results, \citet{Cotta2017}
approached the problem from the point of view of metaheuristics, showing that 
these could outperform the greedy heuristic and tackle problem scenarios
untenable for BnB.

One of the central tenets of these works was a simplistic assumption
on the strategy of the terrorist (whom we should refer to in the following as
\emph{attacker}) whereby the target would be randomly selected. In some
sense, this could capture a total-ignorance scenario in which the defensive
forces (the \emph{defender} in the following) are oblivious to the attacker's 
strategy and so is the attacker with respect to the importance of each target
or the defender's strategy. We here challenge this assumption and consider 
other scenarios in this regard, and how these impact the resulting
optimization task for the greedy heuristic and several metaheuristics. 
Before presenting these algorithms and the experimental setup, 
let us firstly describe in detail the underlying problem, and how the strategy
for decision-taking is established. This is done in next
section.

\section{Definition of the Problem}
\label{sec:background}

As anticipated in the previous section, the main point under scrutiny in this work is the
strategy followed by the attacker in order to select a target. Before describing how we
have modelled the latter, let us firstly describe the basic setting of the model considered.

\subsection{Basic Setting}
\label{sec:setting}
Our goal is to protect a given area under threat, strategically placing detectors in
some points in order to detect and neutralize attackers. To this end, we assume this
area is discretized into a rectangular grid ${A}=\{A_{ij}\}_{m\times n}$. This discretization
allows to determine which portions of the area (to which we shall refer to in the following
as \emph{map}) are walkable (or from a more general point of view, accessible) 
and which ones are inaccessible. The former represent
open spaces through which the attacker can pass and in which we could also place a
detector. As to the latter, they represent any physical obstruction (walls, buildings, street
furniture, etc.) that cannot be traversed by the attacker. Some positions in the map will
also represent \emph{objectives} or targets, that is, sensitive points in which a certain number
of individuals are present and that may be subject to attack. Access to the map is done through
specific positions or \emph{entrances}, which are typically placed in the boundaries of the grid
(but that could in principle be placed anywhere on the grid -- think for example in an
underground passage or metro entrance). Given a certain entrance and a certain
objective, the attacker is assumed to follow a shortest path between these two 
points \citep{Xiaofeng2007,kim16optimal}, using the center of grid cells as
potential waypoints, and avoiding any path intersecting with a \emph{blocked} (i.e., inaccessible) 
cell of the grid -- see \citep{Cotta2017} for a brief discussion on these assumptions. 
In this context, it is easy to construct a weighted graph from a given map, in which vertices
represent grid positions, edges indicate direct accessibility (i.e., the attacker could walk
between these two points in a straight line without intersecting an obstacle), and edge
weights correspond to distances. This graph can in turn be used to compute shortest paths
from entrances to objectives, e.g., using Dijkstra's algorithm or any other suitable method.

\begin{table*}[!t]
\caption{Notation used in the problem modelization. \label{tab:notation}}
\begin{tabular}{lp{.4\textwidth}lp{.4\textwidth}@{}}
\hline
\entrances 	& number of entrances 	& $e_i$	& the $i$-th entrance\\
\objectives 	& number of objectives 	& $o_j$ & the $j$-th objective\\
\detectors 	& number of detectors	& $d_k$	&the $k$-th detector\\
$v$			& speed of the attacker & $t_n$ & time required to neutralize\\
$\eta$ & detector's instantaneous detection rate & \neutral & probability of neutralization \\
$P_{ij}$		& the path between entrance $e_i$ and objective $o_j$ & $\bar{P}_{ij}$ & portion of $P_{ij}$ in which neutralization is still possible.\\
$l_{ijk}$		& portion of $\bar{P}_{ij}$ inside the detection radius of detector $d_k$ & $p_{ijk}$ & probability of $d_k$ detecting the attacker through $\bar{P}_{ij}$\\
$\tilde{D}_{ij}$ & probability of non-detection along $P_{ij}$ & $C_j$ & population at objective $o_j$	\\
$W_{ij}$ & expected number of casualties for $P_{ij}$ & $\gamma_{ij}$ & probability that the attacker picks $P_{ij}$ \\
\hline

\end{tabular}
\end{table*} 

\begin{figure*}[t!] 
\begin{center}
		\includegraphics[angle=0,width=0.5\textwidth]{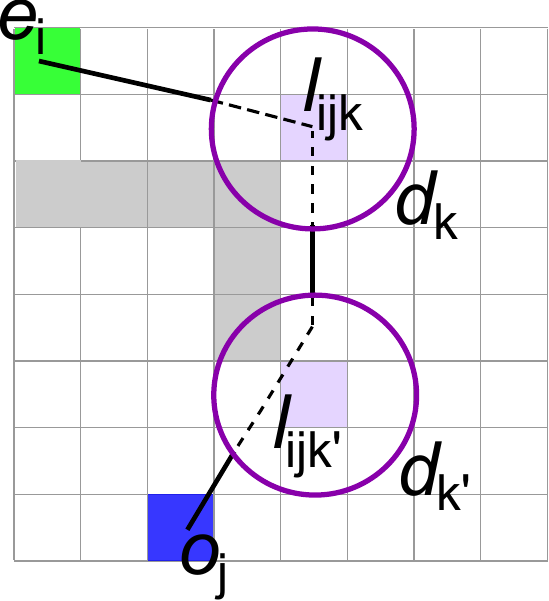}		
		\caption{A shortest path going from entrance $e_i$ to objective $o_j$. Areas monitored by detectors $d_k$ and $d_k'$ are those enclosed by circumferences. The segments of this path detected by each detector are shown with a dotted line and denoted by $l_{ijk}$ and $l_{ijk'}$ respectively.} \label{fig:paths} 
\end{center}
\end{figure*}

As mentioned before, the defense strategy is to place some detectors $d_1, \dots, d_\detectors$
on the map. Each of these detectors is characterized by a detection radius \radius, meaning that 
any attacker path intersecting the circle of radius \radius\ centered at the detector is prone to fire an alarm
that would elicit a neutralization response. Both events, alarm firing and neutralization are not
deterministic in order to capture detector reliability and neutralization effectiveness respectively. 
More precisely, the alarm is fired with a probability that will be larger the longer the section of the path 
included in this detection circle, and the subsequent neutralization attempt will be successful
with some probability \neutral. This can be formalized as follows (see \refTable{tab:notation} for
an overview of the notation used): let $P_{ij}$ be the shortest path
between the $i$-th (out of \entrances) entrance $e_i$ and the $j$-th (out of \objectives) objective $o_j$;
if the attacker was too close to the objective within this path, no neutralization would be possible
(he would reach the target before the enforcing agents could get to him), so let $\bar{P}_{ij}$ be
the portion of the path $P_{ij}$ in which neutralization is still possible (assuming the attacker moves at
speed $v$ and that $t_n$ seconds are required to neutralize him \citep{kaplan05operational}, this amounts 
to detracting a segment of length 
$vt_n$ from the end of  $P_{ij}$). For each detector $d_k$, let $l_{ijk}$ be the portion of  $\bar{P}_{ij}$ 
subject to detection by $d_k$ (see \refFig{fig:paths} for an illustration), an event that will take place with probability 
\begin{equation} 
p_{ijk} = 1-\exp(-\eta l_{ijk})
\end{equation} 
where $\eta>0$ is the detector's instantaneous detection rate. Following \citep{Xiaofeng2007}, detectors 
work independently of one another, and therefore the total
probability of non-detection $\tilde{D}_{ij}$ for path $P_{ij}$ is 
\begin{equation} 
\tilde{D}_{ij} =\prod_{k=1}^\detectors (1-p_{ijk}) = \prod_{k=1}^\detectors \exp(-\eta l_{ijk}) = \exp(-\eta \sum_{k=1}^\detectors l_{ijk})
\end{equation} 
In this case (non-detection) or in case the attacker is detected but not effectively
neutralized (something that will
happen with probability $1-\neutral$), the attacker will reach
the objective $o_j$ causing a number of casualties $C_j$. Thus,
the expected number of
casualties $W_{ij}$ for this particular path will be 
\begin{equation} 
W_{ij} = \tilde{D}_{ij}C_j + (1-\tilde{D}_{ij})(1-\neutral)C_j = C_j\left[\tilde{D}_{ij}\neutral + (1-\neutral)\right] \label{eq:Wij}
\end{equation}
The total expected number of casualties will take into account all
possible paths $P_{ij}$ the attacker can take, that is, between
any entrance $e_i$ and objective $o_j$. To estimate this, let $\gamma_{ij}$
be the probability the attacker chose a particular path $P_{ij}$. Then,
\begin{equation} 
W =\sum_{i=1}^\entrances\sum_{j=1}^\objectives \gamma_{ij}W_{ij}
\label{eq:W}
\end{equation}

Next section will consider different scenarios depending on the particular modelization
of values $\gamma_{ij}$, which capture the strategy used by the attacker to select
a particular course of action.

\subsection{Tackling Decision Making}
\label{sec:decision}
The choice of values $\gamma_{ij}$ can be used to prioritize some targets and paths above others, and hence embodies
the decision-making process done by the attacker. In previous works, i.e., \citep{Xiaofeng2007,Yan201671,Cotta2017},
a very simple approach was considered, namely that the attacker would pick with constant uniform probability any of these paths, that
is, 
\begin{equation}
\gamma_{ij} = \frac{1}{\entrances\objectives}.
\end{equation}
This can be regarded as a completely blind strategy in which the attacker is
fully oblivious to both the damage that can be inflicted by targeting a particular objective or by the potential countermeasures
taken by the defenders, and hence can be arguably considered as unrealistic. To tackle this issue, let us consider more
informed strategies that take the previous points into account.

Let us firstly focus on the importance of each of the objectives. From the point of view of the attacker this is captured
by values $C_j$ that indicate the population at objective $o_j$ (but that in a more general context could be also used to model 
any other feature of objectives whereby their targeting would be valuable for the attacker). An objective-aware strategy would
thus preferably pick objectives of higher value over those with lower value. This can be done in a variety of ways, both
from a qualitative point of view (considering the actual values $C_j$ or only their relative importance as in, e.g., their position
in a ranked list) and from a quantitative point of view (given a certain prioritization index, how these are mapped to actual
probabilities, thus determining how much preference is given to any objective). As a first step, in this work we have considered
a strategy based on using the actual values $C_j$ and defining a proportional selection scheme based on these, i.e.,
\begin{equation}
\gamma_{ij} = \frac{C_j}{\entrances\sum_{k=1}^\objectives C_k}
\end{equation}
Thus, any objective $o_j$ will be targeted with a probability $p_j=\sum_{i=1}^\entrances\gamma_{ij} = C_j/C$, where 
$C=\sum_{k=1}^\objectives C_k$ is the combined value of all objectives, i.e., $p_j \propto C_j$, being all paths to this
objective equiprobable (again, more sophisticated strategies could be thought of here, but we initially assume this for 
simplicity).

Going one step beyond, let us consider awareness by the attacker of the presence of detectors. A game-theory perspective 
would be possible here, by considering that the attacker would select values $\gamma_{ij}$ to maximize his benefit (the
inflicted damage) given whatever knowledge he has about the position of the detectors (or more generally about the strategy
used by the defender to place these), and the defender would in turn place the detectors to minimize his loss given the knowledge 
he has about the attacker strategy. If we define $W[\Gamma,\Delta]$ to be the total number of casualties implied by a particular collection 
$\Gamma$ of values $\gamma_{ij}$ and a particular placement $\Delta$ of the detectors (i.e., a version of \refEq{eq:W} parameterized by $\Gamma$
and $\Delta$), we can see that the solution $(\Gamma^*,\Delta^*)$
given by
\begin{eqnarray}
\Gamma^* & = & \underargmax{\Gamma}{\left(W[\Gamma,\Delta^*]\right)}\label{eq:Gamma} \\
\Delta^* & = & \underargmin{\Delta}{\left(W[\Gamma^*,\Delta]\right)} 
\end{eqnarray}
defines a \emph{Nash equilibrium} state: given a certain location $\Delta^*$ of the detectors, any different choice of values $\gamma_{ij}$
would result is less damage (i.e., a loss for the attacker), and conversely given the attacker's strategy any other placement of the 
detectors would result in larger damage (i.e., a loss for the defender). Notice however that we are adopting here the position of the
defender, and in principle it is not realistic to assume perfect knowledge of the attacker (although we can assume as a worst case scenario that the attacker
has indeed this knowledge about the defender). Thus, if we place detectors at positions given by $\Delta$, we can assume that the attacker would
pick $\Gamma$ to inflict the highest damage as indicated by \refEq{eq:Gamma}. It then follows that we should pick 
\begin{equation}
\Delta^\circ = \underargmin{\Delta}{\left(\undermax{\Gamma}(W[\Gamma,\Delta])\right)} \label{eq:Delta}
\end{equation}
In order to compute the term inside the outer parentheses in the previous equation we could exploit the fact that $W[\Gamma,\Delta]$ is linear
in values $\gamma_{ij}$ and solve a linear program given by
\begin{equation}
\max W[\Gamma,\Delta] = \max \sum_{i=1}^\entrances\sum_{j=1}^\objectives \gamma_{ij}W[\Delta]_{ij}\label{eq:LP}
\end{equation}
subject to
\begin{eqnarray}
\sum_{i=1}^\entrances\sum_{j=1}^\objectives \gamma_{ij} & = & 1\\
0 \leqslant \gamma_{ij}& \leqslant & 1\qquad \forall i\in\{1,\dots,\entrances\}, j\in\{1,\dots,\objectives\}
\end{eqnarray}
where $W[\Delta]_{ij}$ is the value given by \refEq{eq:Wij} parameterized by $\Delta$. We do not need to use a linear programming
solver though, since the solution to \refEq{eq:LP} can be shown to be
\begin{equation}
\gamma_{ij} = \begin{cases}
1 & W[\Delta]_{ij} = \undermax{i'j'}{(W[\Delta]_{i'j'})}\\
0 & {\rm otherwise}
\end{cases}\label{eq:opt}
\end{equation}
i.e., the probability would be concentrated on the path that leads to the largest expected number of casualties (which we shall term
the \emph{critical path}) given the
current distribution $\Delta$ of detectors. 
This follows easily from the fact that $W[\Gamma,\Delta]$ is a convex combination of values $W[\Delta]_{ij}$. Let $i^\circ,j^\circ$ be the indices 
corresponding to the critical path (i.e, $\gamma_{i^\circ j^\circ}=1, \gamma_{ij}=0$ otherwise).
The resulting value of \refEq{eq:LP} for this solution would be $W[\Delta]_{i^\circ j^\circ}$, which by virtue of \refEq{eq:opt} is no smaller
than any other $W[\Delta]_{ij}$, $i,j\neq i^\circ,j^\circ$. Thus, for any other alternative solution with $\gamma_{i^\circ j^\circ}=1-\alpha$, $0<\alpha\leqslant 1$, 
we would have the objective function decreased by $\alpha W[\Delta]_{i^\circ j^\circ}$ and increased at most by
$\alpha W[\Delta]_{i'j'}$ where $W[\Delta]_{i'j'}$ is the second largest expected number of casualties, and hence the resulting value of the objective
function would be reduced. 

Computing the global minimum as per \refEq{eq:Delta} is not trivial, and will be dealt with in next section via heuristic approaches. 
Let us anyway assume that we have found this $\Delta^\circ$, and let $\Gamma[\Delta^\circ]$ be the corresponding path probabilities following
\refEq{eq:opt}. We can see that in purity the state given by $(\Gamma[\Delta^\circ],\Delta^\circ)$ is not a Nash equilibrium point: if $\Delta^\circ$ is
fixed it is certainly the case that the attacker would not benefit from picking different path probabilities as shown before; however, 
taking these path probabilities as fixed, the defender could in general reduce the damage by relocating the detectors so as to have better
coverage of the critical path. This may result in this path no longer being the critical path, hence implying the attacker would pick a
different $\Gamma$; therefore, in general there may not be a Nash equilibrium. At any rate, this point $(\Gamma[\Delta^\circ],\Delta^\circ)$ does indeed capture some kind of equilibrium, were we to
assume that the attacker knows the defender's strategy, namely that the worst-case casualties are to be minimized without taking any risks  
and vice versa, i.e., the
defender knows the attacker will act accordingly to maximize his goals, each one being also aware of their own strategy being known by 
the opponent.

\section{Materials and Methods}
\label{sec:methodology}

In light of the problem setting defined in the previous section, we need to solve
the problem from the perspective of the defender by defining the location
of each of the detectors. This will be done by using different (meta)heuristic approaches
as shown below. In all cases, the objective will be minimizing \refEq{eq:W}, namely 
the expected number of casualties $W[\Gamma,\Delta]$. As shown before, this quantity depends on the
location of the detectors $\Delta$ (which is the actual degree of freedom of the formula, that is,
the part of the equation that will be decided by the optimization algorithm) and the 
path probabilities $\Gamma$ (which will be given by any of the two scenarios depicted before,
that is, fixed and known in advance to be proportional to the importance of each objective,
or variable and known to correspond to the critical path implied by the current location
of detectors). Next we will provide more details about the algorithms considered for
this optimization endeavor, as well as about the benchmark we have used to test these.

\subsection{Algorithmic Approaches}
\label{sec:algorithms}

In this section we describe different algorithmic approaches that we have considered in order to tackle the problem, namely a Greedy Algorithm, a Local Search Procedure, a Tabu Search Metaheuristic and an Evolutionary Algorithm. 

In all cases, the computational cost needed to evaluate a solution can be significantly reduced if some preprocessing is done over the problem instance being considered before starting the optimization process itself. This preprocessing is the same for all the algorithms and thus affects in the same way their performances.
More precisely, a cache memory has been precomputed corresponding to the length of the segment of a path that would be detected in case a detector were placed at a specific cell in the map. These intersecting lengths are independently stored for each possible path going from one of the entrances in the map to one of the objectives. Besides, a dominance memory ${\Delta}=\{\Delta_{ij}\}_{m\times n}$ is also used to prune to some extent the search space that should be explored in order to optimize an specific instance. The implementations of both of these memories are more precisely described in \citep{Cotta2017} and we refer the reader to that work for full details.

\subsection{Greedy Algorithm}
\label{sect:GREEDY}
The algorithm described in this section (denoted as \Greedy{}) corresponds to the one originally proposed by \citet{Xiaofeng2007}. This algorithm is a constructive method hence it places on the map the \detectors\ 
detectors one after the other. The extension of the current partial solution is done by adding on each step the detector which leads to a better local extension.

\begin{algorithm}[!t]
	\caption{Greedy Algorithm\label{fig:alg:GREEDY}}
	\KwIn{\detectors\ (number of detectors to be placed)}	
	$\mathit{candidates} := \{(r,c)\ |\ 1 \leqslant r \leqslant m, 1 
	\leqslant c \leqslant n, A_{rc}$ is unblocked $, 
	\Delta_{r,c}<\detectors \} $\;	
	$\mathit{sol} := \varnothing$\;
	\While{$\mathit{length}(\mathit{sol}) < \detectors$}
	{	
		$\mathit{fitness}^* := \infty$\;
		$\mathit{detector}^* := \mathrm{null}$\;
		\For {$d \in \mathit{candidates}$}
		{
			$\mathit{tentative} := \mathit{sol} + \{ d \}$\;
			\If{$\mathit{value}(\mathit{tentative}) < \mathit{fitness}^*$}
			{
				$\mathit{detector}^* := d$\;
				$\mathit{fitness}^* := \mathit{value}(\mathit{tentative})$\;
			}		 
		}
		$\mathit{sol} := \mathit{sol} + \{\mathit{detector}^* \}$\;
		$\mathit{candidates} := \mathit{candidates} \setminus 
		\{\mathit{detector}^*\}$\;
	}	
	\Return{$\mathit{sol}$}\;
\end{algorithm}

The corresponding pseudo-code is shown in \refAlg{fig:alg:GREEDY}. 
Firstly, all non-blocked cells in the 
map minus those that are dominated by at least \detectors\ cells are taken as potential detector placements (line 1).
Then, starting from an empty solution (line 2), the \detectors\ detectors 
are added to the current partial solution ($\mathit{sol}$) one at a time (lines 3-15). 
On each iteration, the candidate detector whose 
incorporation leads to the best extended solution --i.e., the one locally minimizing the total number of casualties given the current location of detectors and the attacker model considered-- is selected (lines 6-12), added to the solution (line 13) and removed from the set of candidate detectors (line 14).

As a greedy algorithm, this procedure makes different locally optimal decisions  and
 hence the resulting final solution cannot be guaranteed to result optimal.

\subsection{Hill Climbing}
\label{sect:HC}
We also consider a Hill Climbing (HC) Local Search algorithm. In contrast to constructive methods --such as the \Greedy{} method described above-- which work with a partial solution, these algorithms work with a complete solution during the optimization process \citep{AartsLenstra97,HoosStutzleBook}. A crucial concept here is that of {\em neighborhood} which refers to the set of solutions which can be obtained from current one by modifying one part of the solution. The algorithm starts from a complete solution that should be constructed somehow. Subsequently, solutions in the neighborhood of the current one are examined and evaluated and a move towards one of these is performed in case the quality of such solution improves the current one. This process will be repeated until no solution in the neighborhood of the current one provides any improvement. At this point, a local optimum to the problem has been found and is returned. The whole process can be iterated starting from different initial solutions until the allowed execution time is exhausted, returning the best found solution at the end.

The pseudo-code for this procedure is shown in \refAlg{fig:alg:HC}. $sol$ stands for a complete solution, implemented as a 
vector of different locations for placing the detectors. The algorithm explores the neighborhood of the current solution by substituting one detector location 
in the current solution by each of the unused candidate locations for the problem instance 
($\fnctn{Replace}(\mathit{sol},i,d)$ denotes the solution that is obtained by replacing the $i$-th 
detector in solution $\mathit{sol}$ by the alternative detector $d$ -- see line 7). Then the best modified solution is selected (lines 8-11). If this selected solution is better than the current one, a move is done towards the new solution (line 9) and an improvement has been achieved (line 10). Starting from this improved solution, the same procedure is done for the remaining detectors, and this process is repeated until a local optimum is reached, i.e., when no further improvement is possible.

\begin{algorithm}[!t]
	\caption{Hill Climbing Algorithm	\label{fig:alg:HC}}
	\KwIn{$\mathit{sol}$ (a collection with the coordinates of \detectors\ detectors)}	
	$\mathit{candidates} := \{(r,c)\ |\ 1 \leqslant r \leqslant m, 1 \leqslant c \leqslant n, A_{rc}$ is unblocked$, \Delta_{rc}<\detectors \} $\;
	$\mathit{improvement} := \mathrm{true}$\;
	\While{$\mathrm{improvement}$}
	{
		$\mathit{improvement} := \mathrm{false}$\;
		\For{$i := 1$ \kw{to} $\mathit{length}(\mathit{sol})$}
		{
			\For{$d \in (\mathit{candidates} \setminus \{d\,|\,d \in \mathit{sol} \})$}
			{		
				$\mathit{tentative} := 
				\fnctn{Replace}(\mathit{sol},i,d)$\;
				\If{$\mathit{value}(\mathit{tentative}) < \mathit{value}(\mathit{sol})$}
				{		
					$\mathit{sol} := \mathit{tentative}$\;
					$\mathit{improvement} := \mathrm{true}$\;
				}
			}
		}
	}	
	\Return{$\mathit{sol}$}\;
\end{algorithm}

\subsection{Tabu Search}
\label{sect:TS}

Tabu Search (TS) is a metaheuristic
originally proposed by \cite{gloverTS89,gloverTS90} that has been
successfully used for the resolution of different combinatorial optimization problems. The key idea in TS is to use a memory data structure in order to keep track of the previous history of the search. This memory is referred to as 
the {\em Tabu List}, as moves towards solutions included in this list are
forbidden. This memory serves the purpose of avoiding
revisiting recently explored solutions, thus allowing the algorithm to avoid to a certain extent the possibility of getting trapped in cycles during the search process. At each iteration, the search moves towards the best solution in the neighborhood of allowed moves. Since all improving solutions may be forbidden, moves towards solutions worse than the current one may be performed and this is the mechanism allowing the algorithm to escape
from local optima. As keeping all visited solutions in a list is not memory efficient, an usual alternative consists of using a data
structure keeping track of attributes of  recently visited solutions, and forbidding moves towards solutions with those same
attributes for some iterations (the so called {\em tabu tenure}). 

Our implementation of a TS algorithm is shown in \refAlg{fig:alg:TS}. A solution corresponds to a vector with \detectors\ locations. The neighborhood of a solution consists of all solutions that can be obtained from that one by moving one of its detectors to another location not included in the solution. As tabu structure, we use a matrix ${T}=\{T_{ij}\}_{m\times n}$ so that $T_{ij}$ records tabu information for doing a move consisting in placing a detector at cell $i \times j$ in the map. $\mathit{sol}$ is the solution currently being explored and $\mathit{best}$ is the best solution found so far. The algorithm repeats the following process, with a limit on the number of iterations to perform in case no improved solution is attained:
\begin{itemize}
	\item All moves leading to solutions in the neighborhood of the current one which are not currently tabu are considered and those producing solutions with the best value are collected into $\mathit{bestMoves}$ set (lines 5-18). Additionally, if one solution in the neighborhood is better than the best one found so far (the so named {\em aspiration criteria}), this solution is also considered even if the corresponding move were forbidden by the tabu memory.
	\item Then, one specific move among the best ones is randomly chosen (line 19) and performed (line 20). Recall that a move displaces one detector to a new position. Trying to move one detector to this same position is set as tabu for a number of iterations (line 21).
	\item Finally, if the new solution is the best one found so far, it is recorded and an improvement is acknowledged (lines 23-24). Otherwise, we record that this iteration was unable to achieve any improvement (line 26).   
\end{itemize}

Just like in the case of \HC{} algorithm, the whole process is repeated several times until the allowed execution time is exhausted, and the best solution found is returned at the end.

\begin{algorithm}[!t]
	\caption{Tabu Search Algorithm	\label{fig:alg:TS}}
	\KwIn{$\mathit{sol}$ (a collection with the coordinates of \detectors\ detectors)}	
	$\mathit{candidates} := \{(r,c)\ |\ 1 \leqslant r \leqslant m, 1 \leqslant c \leqslant n, A_{rc}$ is unblocked$, \Delta_{rc}<\detectors \} $\;
	$\mathit{best} := \mathit{sol}$\;
	$\mathit{noImprovement} := 0$\;
	\While{$\mathit{noImprovement} < \mathit{maxIters}$}
	{		
		$\mathit{bestMoves} := \varnothing$\;
		$\mathit{bestMovesValue} := \infty$\;
		\For{$i := 1$ \kw{to} $\mathit{length}(\mathit{sol})$}
		{
			\For{$(r,c) \in (\mathit{candidates} \setminus \{d\,|\,d \in \mathit{sol} \})$}
			{		
				$\mathit{tentative} := \fnctn{Replace}(\mathit{sol},i,(r,c))$\;
				\If{$\kw{not}\ \fnctn{Tabu}(T,r,c)\ \kw{or}\ \mathit{value}(\mathit{tentative}) < \mathit{value}(\mathit{best})$}
				{
					\uIf{$\mathit{value}(\mathit{tentative}) = \mathit{bestMovesValue}$}
					{		
						$\mathit{bestMoves} := \mathit{bestMoves} + \{\mathit{(i,r,c)}\}$\;
					}
					\uElseIf{$\mathit{value}(\mathit{tentative}) < \mathit{bestMovesValue}$}
					{
						$\mathit{bestMoves} := \{\mathit{(i,r,c)}\}$\;
						$\mathit{bestMovesValue} := \mathit{value}(\mathit{tentative})$\;
					} 
				}
			}
		}
		$\mathit{(i,r,c)} := \fnctn{chooseFrom}(\mathit{bestMoves})$\;
		$\mathit{sol} := \fnctn{Replace}(\mathit{sol},i,(r,c))$\;
		$\fnctn{setTabu}(T,r,c)$\;	
		\uIf{$\mathit{value}(\mathit{sol}) < \mathit{value}(\mathit{best})$}
		{
			$\mathit{best} := \mathit{sol}$\;
			$\mathit{noImprovement} := 0$	
		}
		\Else
		{
			$\mathit{noImprovement} := \mathit{noImprovement} + 1$
		}
	}	
	\Return{$\mathit{best}$}\;
\end{algorithm}

\subsection{Evolutionary Algorithm}
\label{sect:EA}
An Evolutionary Algorithm (EA) \citep{Eiben:2003:IEC} is a optimization 
procedure inspired by the biological evolution of species. These algorithms are population based ones which means that a pool of solutions is maintained during the optimization process. This population of solutions is initialized somehow before entering the evolutionary loop in which solutions go through different processes such as reproduction, recombination, mutation and replacement. At the end of the evolutionary loop, the best solution found during the execution of the algorithm is returned.

\refAlg{fig:alg:EA} shows the specific incarnation of the EA that we have used. 
$\mathit{pop}$ is the population of non-repeated $\mathit{popSize}$ 
individuals, each one being a full solution to the problem. Like in previous heuristics, the search space explored by the EA 
is restricted to non-blocked cells in the 
map minus those that are dominated by at least \detectors\ cells.
Each individual in the population of solutions is initialized randomly as a vector of \detectors\ different locations for placing each detector. These locations are randomly chosen in an uniform way among all possible ones (lines 1-4). Then, the following process corresponding to the evolutionary loop is repeated, until the  maximum allowed execution time for the algorithm is exhausted:
\begin{itemize}
	\item With probability of selection $p_{X}$, two individuals acting as parents are selected from the population (lines 7-9). The specific algorithm used for each selection is  {\em binary tournament selection} in which, after picking two individuals at random, the best one is selected. Now, these individuals go through recombination in order to produce a new one. For this purpose, let 
	us consider the set comprising the union of detector placements 
	included in to be recombined individuals. A new individual, constituting the offspring for this 
	generation, is then defined by sampling in a uniform way \detectors{} 
	elements (i.e., detector placements) from this previous set.
	\item Otherwise (i.e. with probability $1-p_{X}$) recombination is not performed and a random individual of current 
	population is selected as the offspring (line 11).
	\item Each detector location in the the offspring is mutated with probability $p_m$. More precisely, with such probability the corresponding detector is replaced with another one not 
	included in the solution (line 13). The resulting individual is evaluated (line 14).
	\item As to replacement, the worst individual currently included in the  population is replaced by the offspring (line 15). 
\end{itemize}
Finally, when the execution time limit is reached, the best individual found along this whole process is returned as a solution.

\begin{algorithm}[!t]
	\caption{Evolutionary Algorithm\label{fig:alg:EA}}
	\SetKwInOut{Input}{Input}  
	\Input{$\mathit{popSize}$ (population size)\\
		$p_{X}$ (recombination probability)\\
		$p_m$ (mutation probability)}	
	\For{$i :=1$ \kw{to} $\mathit{popSize}$}
	{
		$\mathit{pop}_i :=$ \fnctn{RandomIndividual}$()$\;
		\fnctn{Evaluate}($\mathit{pop}_i$)\;
	}
	\While{allowed runtime \kw{not} exceeded}
	{
		\eIf{recombination is performed($p_{X}$)}
		{
			$parent_1 :=$ \fnctn{Select}($\mathit{pop}$)\;
			$parent_2 :=$ \fnctn{Select}($\mathit{pop}$)\;
			$\mathit{offspring} :=$ \fnctn{Recombine}($parent_1$, $parent_2$)\;
		}{
			$\mathit{offspring}:=$ \fnctn{Select}($\mathit{pop}$)\;
		}
		$\mathit{offspring} :=$ \fnctn{Mutate}($p_m,\mathit{offspring}$)\;
		\fnctn{Evaluate}($\mathit{offspring}$)\;
		$\mathit{pop} :=$ \fnctn{Replace}($\mathit{pop}$, $\mathit{offspring}$)\;
	}	
	\Return{best solution found}\;
\end{algorithm}

\subsection{Problem Instances}
\label{sec:instances}

\begin{figure*}
\subfloat[\label{fig:map:example:harbour}]{\includegraphics[width=.325\textwidth]{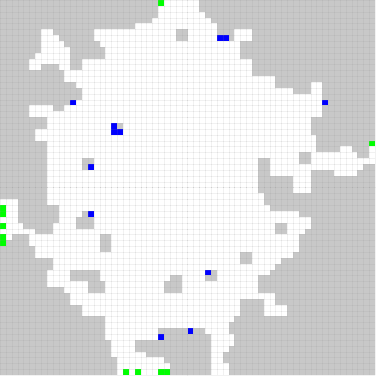}}~
\subfloat[\label{fig:map:example:newtown}]{\includegraphics[width=.325\textwidth]{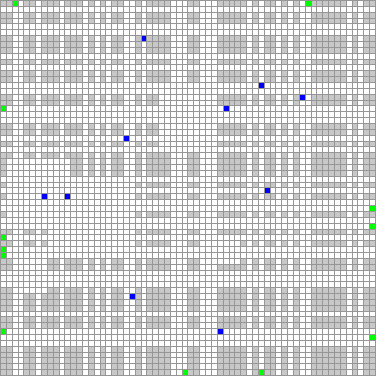}}~
\subfloat[\label{fig:map:example:oldtown}]{\includegraphics[width=.325\textwidth]{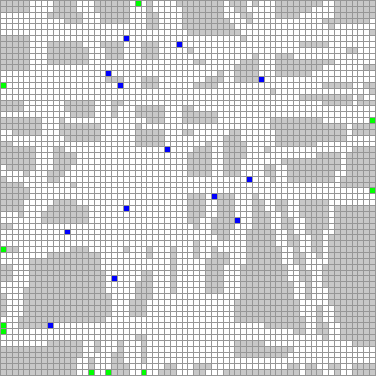}}
\caption{Examples of each of the three scenarios considered. (a) \harbour\ (b) \newtown\ (c) \oldtown.}\label{fig:map:example}
\end{figure*}

In order to test the algorithms described in the previous section in a broad set of conditions,
we have generated problem instances that try to resemble different real-world scenarios. To be precise,
we have considered the following three classes of maps:
\begin{itemize}
	\item \harbour: 	these problem instances represent a coastal area that may be subject to a 
					maritime attack by a small vessel, cf. \citep{Yan201671}. These maps feature
					a large inner open area (representing a water mass such as a lake or a
					sheltered body of sea) typically comprising several scattered islands, 
					surrounded by a complex coastline with some straits from which the attacker
					can enter the threat area. See \refFig{fig:map:example:harbour}. 
	\item \newtown: these problem instances represent a modern urban area in which streets run
					in straight lines creating a grid (Manhattan or Barcelona's Expansion District being
					prominent examples). These maps feature streets of different width running at
					right angles, as well as a number of plazas of different sizes scattered across
					the map.  See \refFig{fig:map:example:newtown}. 
	\item \oldtown: 	these problem instances represent the historic core of an old town. As such, 
					the street plan features a much more decentralized and flexible structure. Streets 
					typically originate at plazas and run in different directions and angles with 
					respect to each other, bifurcating as they get farther from the plazas.  See \refFig{fig:map:example:oldtown}. 
\end{itemize}

For each of these three problem classes, we have defined
a map generator in order to create a number of instances of the desired characteristics, hence contributing to the 
diversity of the benchmark. Next we will describe in more detail the generation procedure for each map class.

\subsubsection{Coastal Area}
These maps are created in two phases. In the first phase a exponentially decaying diffusion mechanism is used to grow the
inner water mass: we start at the center of the lattice with a value $p=1$; with this probability we set the current position to 
unblocked and move to each of the four neighbors in a von Neumann topology (i.e., North, East, South, West), having this
probability decay by a different factor $\{\beta_i\}_{1\leqslant i\leqslant 4}$, that depends on the direction, and repeating the
process from each of them. When the diffusion stops, we proceed with the second phase in which the map is smoothed out 
by using a cellular automaton which has each lattice cell change to the most repeated state among itself and its neighbors.
This process is run until convergence or after a maximum number of iterations has been performed. 

Once the above is done, objectives are placed on positions along the coastline. To do so, a random unblocked position is
selected, and a random walk (using the von Neumann connectivity mentioned before) is performed until a blocked position
is found. This position is then unblocked and defined as an objective. The process is repeated as many times as objectives
need to be placed.  The entrances are randomly selected among the unblocked positions on the borders of the map.

\subsubsection{Modern Urban Area}
These maps are also created in two phases: in the first one the street grid is created, and in the second one a  number
of plazas are added to the map. In order to tackle the first part, we consider values $\{p_i\}_{0\leqslant i}$, representing
the probability of having a street of width $i$ (the value $p_0$ corresponding to the situation of not having a street at
all). Using these, we traverse the columns of the maps deciding whether we include a North-South street and if so of which 
width. In case a street of width $w$ is placed, we jump $w+1$ columns to the right and repeat the procedure until all
columns have been processed. The whole process is then repeated by traversing rows of the maps in order to place
West-East streets.

The second phase involves placing some plazas on the map. This is controlled by parameter $n_p$ indicating the
number of plazas and values $\{q_i\}_{\min_p\leqslant i}$ representing the probability of having a square plaza of a certain
side length, where $\min_p$ indicates the minimum side length admissible. The generator would iterate $n_p$ times
a process of determining the size of a plaza using probabilities $\{q_i\}$ and then deciding a suitable location given this size. 
As in the previous scenario, the entrances are randomly selected among the unblocked positions on the borders of the map.
As to the objectives, these are placed in random unblocked positions of the map.

\subsubsection{Old Historic Town}
These maps are created by iterating $n_p$ (a parameter that just like in the previous scenario indicates a number
of plazas to be located on the map) times a basic procedure. This procedure consists of determining the size of a plaza
(using a collection of values $\{q_i\}_{\min_p\leqslant i}$ as in the \newtown\ scenario), placing it in on the map, and then
having four streets depart from each of the sides of the plaza. These streets do not run at right angles though: they 
are laid out having a certain slope between $\pm 1$ (a diagonal street) and $\pm \max_s$ (closer to horizontal or vertical).
The width of the street is set to $w$ and as the street runs out from the plaza it can branch along the way into a smaller (width $w-1$)
street with probability $p_b$.  These branched-out streets can in turn branch into smaller lanes in a recursive way, thus 
trying to resemble the complex and more convoluted network of small streets and lanes in the historic core of old towns.

Much like it was done in \harbour\ and \newtown, the entrances are randomly selected among the unblocked positions 
on the borders of the map. As to the objectives, these are placed in random unblocked positions of the map as in \newtown.

\section{Experimental Results}
\label{sec:results}

In order to evaluate the performance of the algorithms we have followed
a problem generator approach \citep{DBLP:conf/icga/JongPS97}, whereby
rather than running each algorithm multiple times on the same problem instance, they
are run once on a different problem instance each time, thus allowing to obtain a more 
representative measure of performance across the whole set of possible instances, 
and avoiding (or at the very least diminishing the impact of) any spurious match 
between a specific instance and a specific algorithm. More precisely, all algorithms
have been run twenty five times on each of the three problem classes, 
namely \harbour,
\newtown\ and \oldtown, for the two
attacker scenarios (i.e., proportional selection of objectives and worst-case equilibrium)
using the same seventy five problem instances in all cases.

The maps considered have been generated on a grid of size $64\times 64$. In all
cases, the number of entrances \entrances\ and the number of objectives \objectives\ 
are independently chosen uniformly at random between 10 and 15. A 10\% region around
the borders of the maps is kept free of objectives to avoid any of these being located
extremely close to an entrance, something which could be regarded as a pathological case.
As to scenario 
dependent parameters, in the case of \harbour, the decay parameters $\{\beta_i\}$ 
were randomly picked between 0.98 and 0.99 and checked to empirically ensure that 
the diffusion barely reached the borders of the map.
The dimension of each cell in the map was of 200m $\times$ 200m and the detection radius for the detectors was set to $\tau=500m$. Regarding \newtown, each map has a
random number of plazas $n_p$ between 3 and 6, whose side length can range between 4 and 
13 ($\{q_i\}_{4\dots13}=1/10$). Street widths are selected using values $\{p_i\}_{0\dots3}=\{0.5, 0.25, 0.15, 0.1\}$. Finally,
\oldtown\ maps also comprise a random number of plazas $n_p$ between 3 and 6, whose
sizes range from 6 up to 15 ($\{q_i\}_{6\dots15}=1/10$). The maximum slope of streets is 1/5 of the map side length,
they have initial width $w=3$, and can branch with probability $p_b=0.02$. For these instances representing different town configurations, the dimension of each cell in the map was of 5m $\times$ 5m and the detection radius for the detectors was set to $\tau=20m$.

We have considered that minimum time required to neutralize an attacker is $t_n=10$s. In the case of \newtown\ and \oldtown\ instances, this means that effective neutralization is not possible if the distance of the attacker to the objective is less than 10m, which corresponds to assuming that the attacker moves at a speed $v=1$m/s. In the case of \harbour\ instances, we have assumed a moving speed for the vessels $v=20$m/s and thus the maximum distance to the objective for being able to abort the attack is 200m.
As to the detector's instantaneous detection rates, we have used a value of $\eta=0.06$ for the \newtown\ and \oldtown\ instances and $\eta=0.006$ for 
the \harbour\ instances, thus assuming that detectors are less reliable in the later case.
For the probability of effectively neutralizing a detected attacker, we used $\neutral=0.6$ in all cases. Additionally, in order to estimate the expected number of casualties for an objective cell $\{C_{j}\}_{1\dots\objectives}$, we used the same procedure defined by \citet{Cotta2017} which is in turn based on
Equation 2 in \citep{kaplan05operational}. This equation presumes that the number of fragments produced after the explosion tends to $\infty$ and that these fragments and individuals around the target area are spatially distributed according to a Poisson process. We have used the same parameters for this equation as those proposed by \citet{Xiaofeng2007}, except for the population densities near each objective cell, which were set as a constant therein but which we have modelled as a random variable instead, with the aim of considering more diverse instances.  For \newtown\  and \oldtown\ instances this
is chosen from a normal distribution ${\mathcal N(0.4, 0.1)}$ persons / ${\mathrm m}^2$, whereas in the case of \harbour\ instances, we take ${\mathcal N(9 \cdot 10^7, 1.8\cdot 10^6)}$ (these latter values do not really stand for an expected number of civilian casualties but instead represent the economical cost resulting from damaging the corresponding objective).
Overall, all parameters have been picked to have generated instances resemble 
realistic scenarios, as described in the literature \citep{kaplan05operational,Xiaofeng2007,Yan201671}, while providing diversity thanks to their range of variability.
%

Regarding the parameters used for the different algorithms, the following settings were used for the EA: population size $\mathit{popSize}=100$ individuals, probability of crossover $p_X=0.9$ and probability of mutation $p_m=1/\detectors$.
The {\em tenure value} for the TS algorithm was a random number of iterations in $[\detectors \dots 2\detectors]$. All algorithms
were run for 30 seconds on each instance (the hardware platform used was a cluster of Intel Xeon E7-4870 2.4 GHz processors with 2GB of RAM running under SUSE Linux Enterprise Server 11 operating system).

\begin{figure*}[!t]
\centerline{\includegraphics[width=\textwidth]{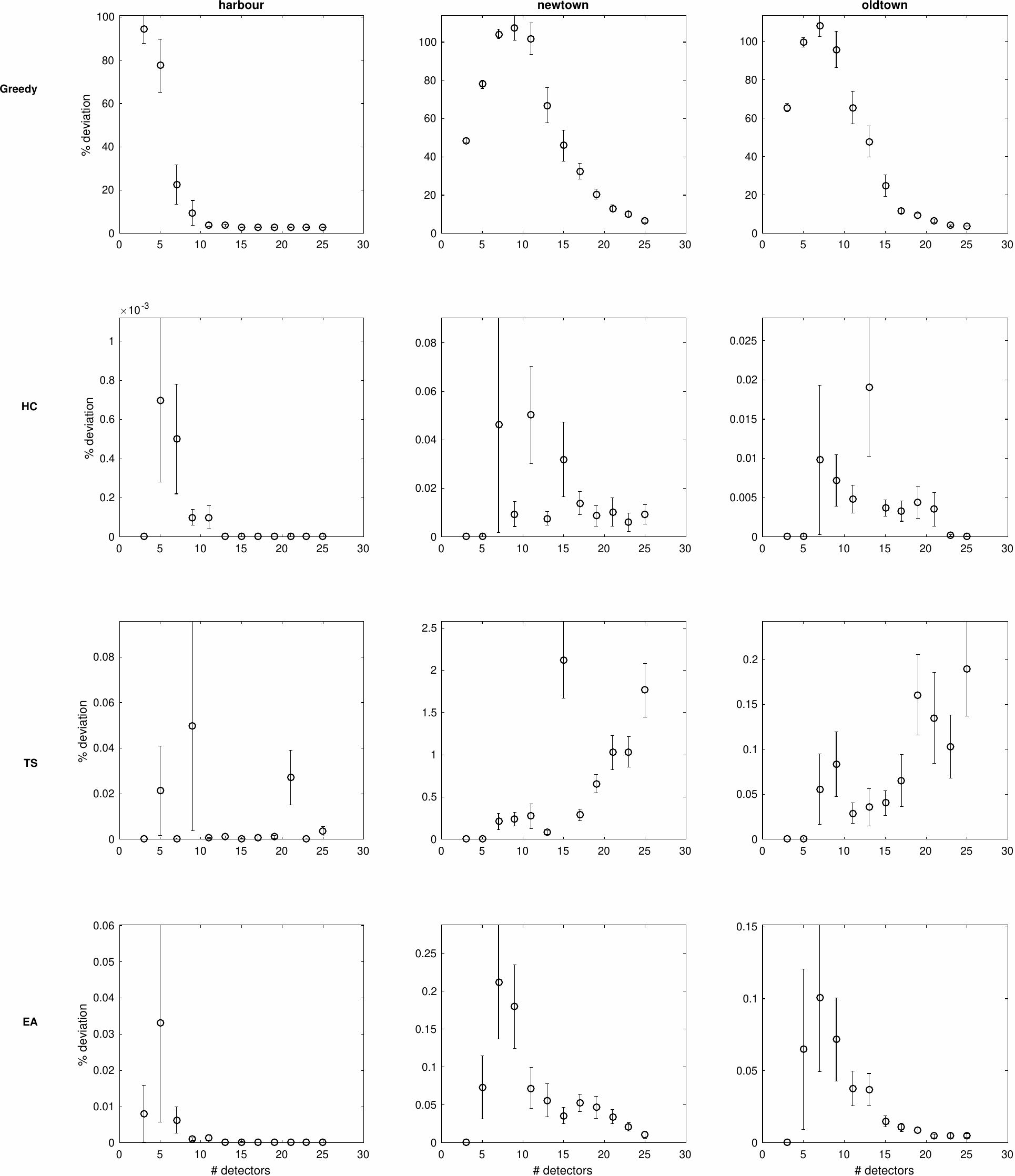}}
\caption{Deviation (\%) from the best known solution for each algorithm and data set as a function 
of the number of detectors used in the proportional selection scenario. Notice the different scale
on the Y axis in each subfigure. \label{fig:errorbar:prop}}
\end{figure*}

\begin{figure*}[!t]
\centerline{\includegraphics[width=\textwidth]{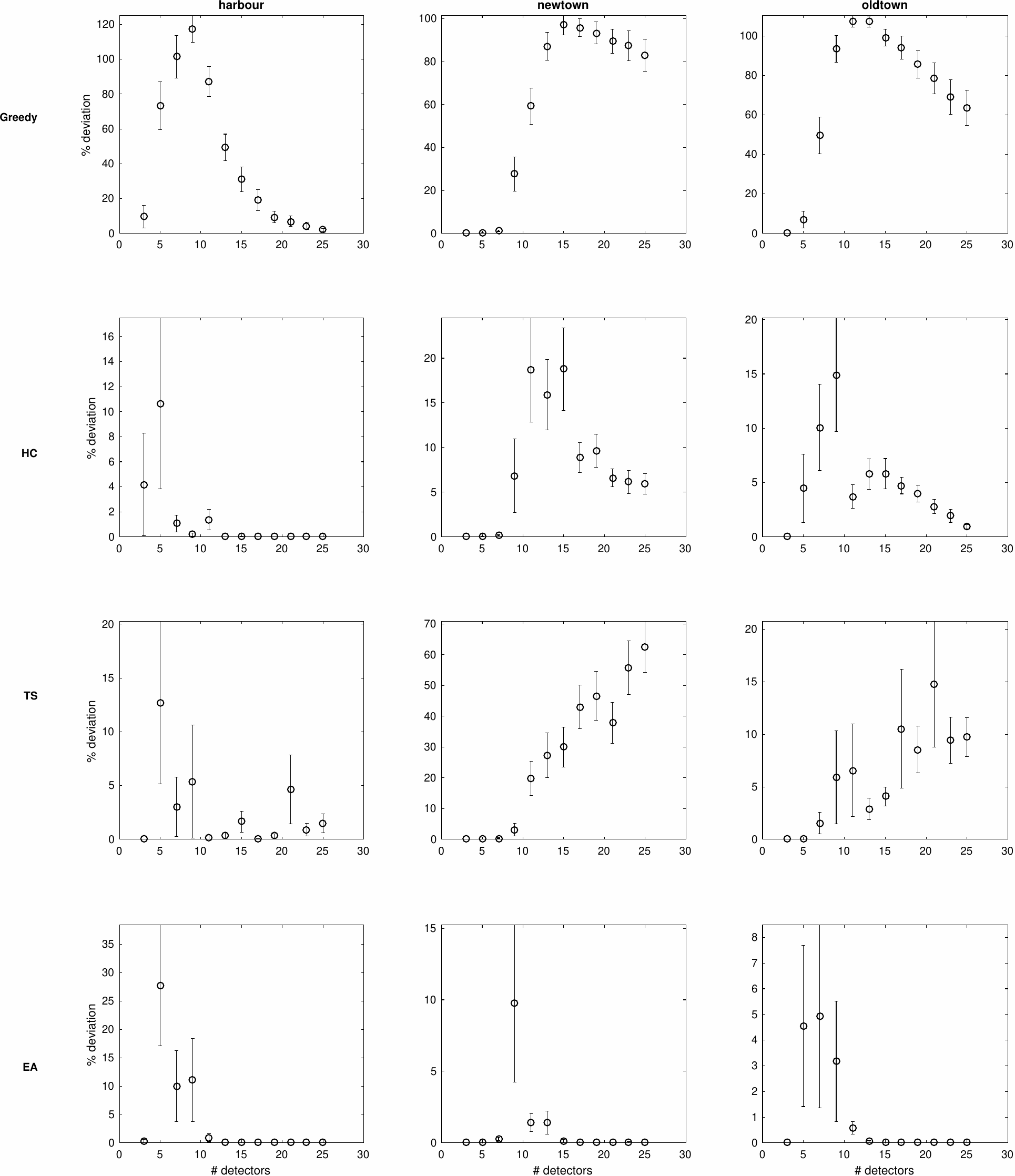}}
\caption{Deviation (\%) from the best known solution for each algorithm and data set as a function 
of the number of detectors used in the worst-case equilibrium scenario. Notice the different scale
on the Y axis in each subfigure. \label{fig:errorbar:nash}}
\end{figure*}

The results are depicted in \refFigs{fig:errorbar:prop}{fig:errorbar:nash}. In the two scenarios,
the three iterative approaches largely outperform the greedy algorithm. In most cases, the latter
seems to experience a peak of difficulty when the number of detectors is about the same magnitude
as the number of objectives. This can be explained by the fact that for a low number of detectors 
the number of options are limited and the greedy selection is often reasonably close to more finely
adjusted solutions, whereas for a large number of detectors these can be spread to cover most paths
without not so much difficulty (thus leaving the middle regime between these two extremes
as the most complex one). 

\begin{figure*}[!t]
\subfloat[\label{fig:boxplot:prop}]{\includegraphics[width=.5\textwidth]{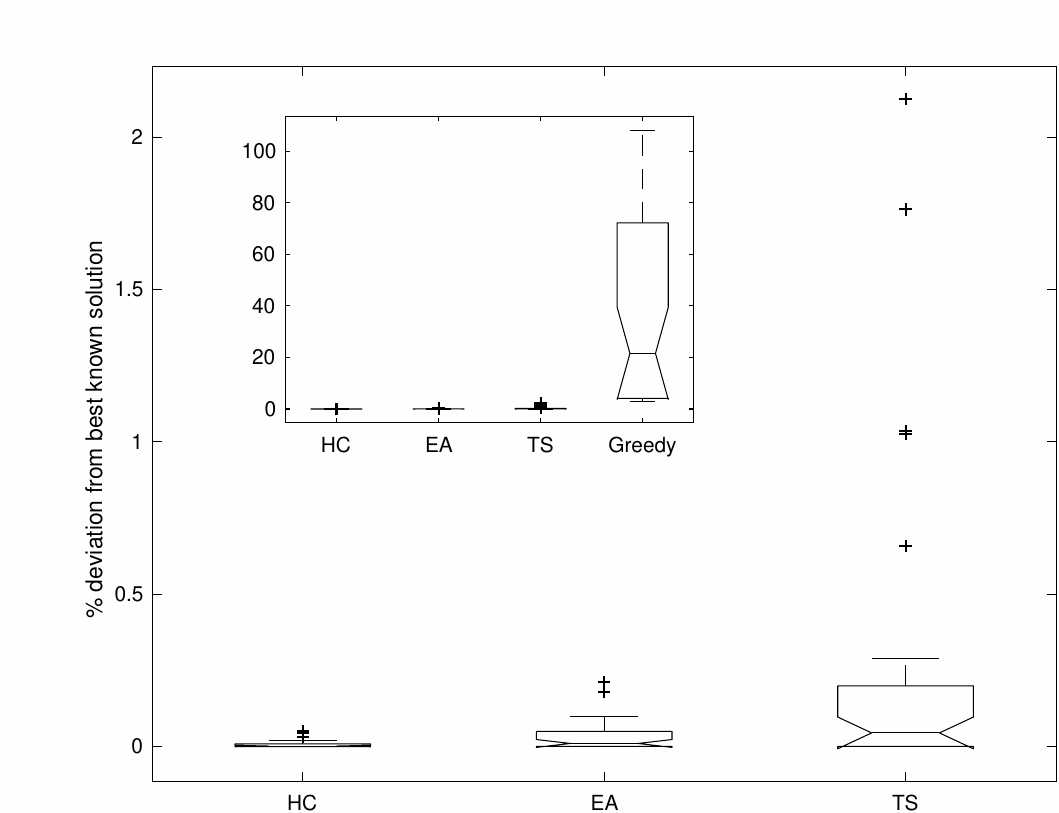}}
\subfloat[\label{fig:boxplot:nash}]{\includegraphics[width=.5\textwidth]{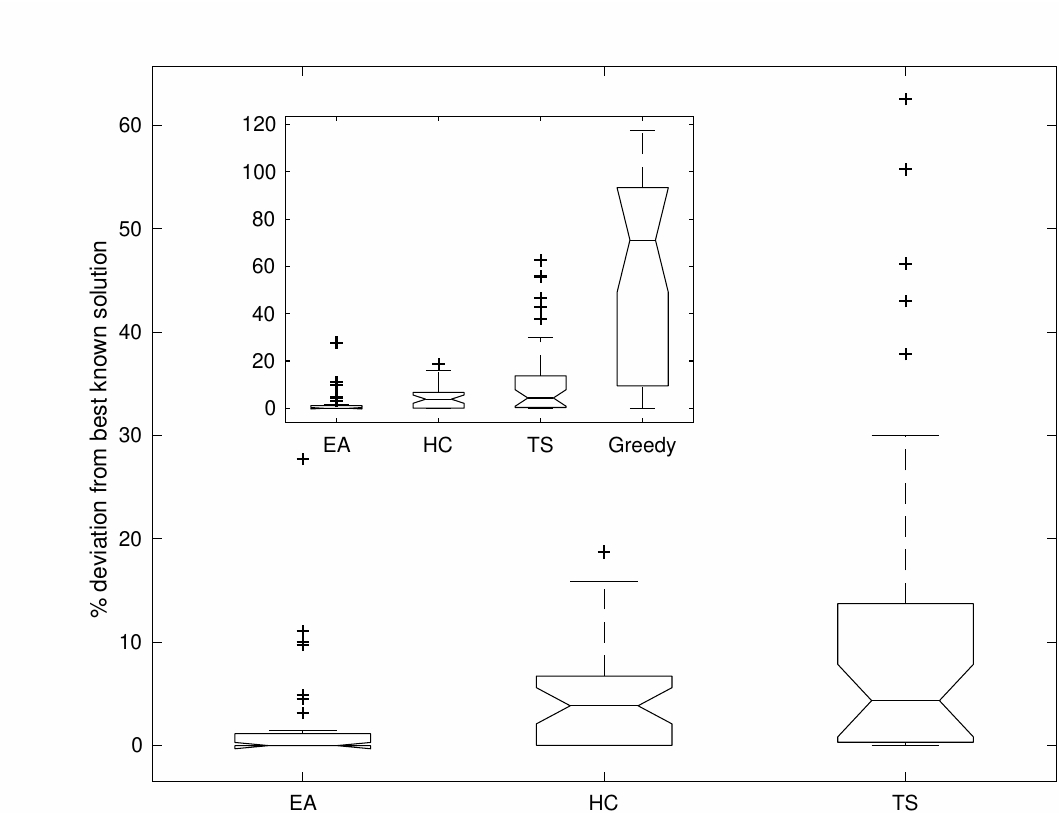}}
\caption{Cumulative boxplot of deviations (\%) from the best know solution for each algorithm. The main
plot focus on the iterative heuristics, and the inset includes the greedy algorithm as well. (a)
Proportional selection (b) Worst-case equilibrium. Notice the different scale
on the Y axis in each subfigure.\label{fig:boxplot}}
\end{figure*}

This said, it is much more interesting to note the difference between the two scenarios. Notice firstly
how the range of deviations for all algorithms is larger in the worst-case equilibrium scenario. It is
not difficult to see that this latter scenario is indeed harder from an optimization point of view, due to 
the complex structure of the search landscape in this case: the quality of a solution is dictated by the
critical path given the current location of detectors; hence, the optimization algorithm will attempt
to gradually increase the coverage of this critical path but in doing so (i) it will have to take care of
not leaving uncovered other paths to objectives of higher sensitivity, and (ii) eventually the critical
path may turn to be a different one (recall the discussion in \refSect{sec:decision}) causing an
abrupt discontinuity in the direction of the optimization process. This differs from the proportional
selection scenario, in which the objective function is a weighted combination whose coefficients
are fixed, and hence can be optimized in a more smoothly way. Furthermore, notice how the search
landscape is prone to have mesas in the worst-case equilibrium scenario, given the fact that as long
as the coverage of the critical path does not change and the coverage of other paths does not change
enough for the critical path to be a different one, the value of the objective function will be the same.
This may affect all algorithms, and in particular the greedy algorithm (see the insets in \refFig{fig:boxplot}), 
which will face the need to pick a choice among a large number of seemingly equivalent options.

\begin{table*}[!t]
\caption{Results of Holm Test in the proportional selection scenario using \HC\ as control algorithm. \label{tab:holm_prop}}
\begin{tabular}{lllrrcr}
\hline
$i$ & ~\hspace{2mm}~& algorithm         & ~\hspace{2mm}~$z$-statistic & ~\hspace{2mm}~$p$-value & ~\hspace{1mm}~adjusted $p$-value & \\
\hline	
1 && EA		& 3.38e$+$00 &	3.66e$-$04	&	3.66e$-$04		\\	
2 && TS		& 4.56e$+$00 &	2.51e$-$06	&	5.01e$-$06		\\	
3 && Greedy	& 9.22e$+$00 &	1.49e$-$20	&	4.46e$-$20		\\	
\hline
\end{tabular}
\end{table*}

\begin{table*}[!t]
\caption{Results of Holm Test in the worst-case equilibrium scenario using the \EA\ as control algorithm. \label{tab:holm_nash}}
\begin{tabular}{lllrrcr}
\hline
$i$ & ~\hspace{2mm}~& algorithm         & ~\hspace{2mm}~$z$-statistic & ~\hspace{2mm}~$p$-value & ~\hspace{1mm}~adjusted $p$-value & \\
\hline
1 && HC		& 1.64e$+$00 & 	5.02e$-$02	& 5.02e$-$02		\\
2 && TS		& 3.29e$+$00 & 	5.08e$-$04	& 1.02e$-$03		\\	
3 && Greedy	& 8.22e$+$00 &	1.05e$-$16	& 3.16e$-$16		\\
\hline
\end{tabular}
\end{table*}

\begin{figure*}[!t]
\subfloat[\label{fig:newtown:nash:greedy}]{\includegraphics[width=.5\textwidth]{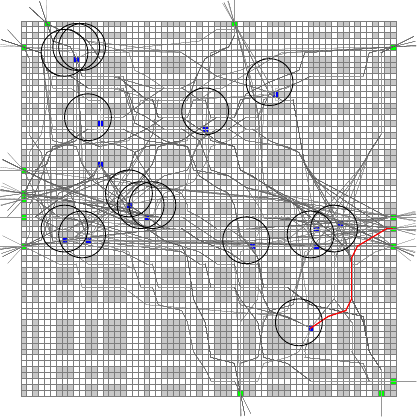}}
\subfloat[\label{fig:newtown:nash:hc}]{\includegraphics[width=.5\textwidth]{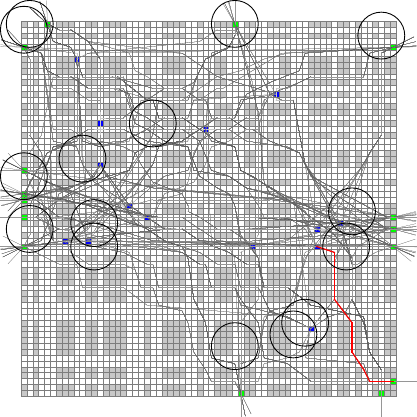}}

\subfloat[\label{fig:newtown:nash:rts}]{\includegraphics[width=.5\textwidth]{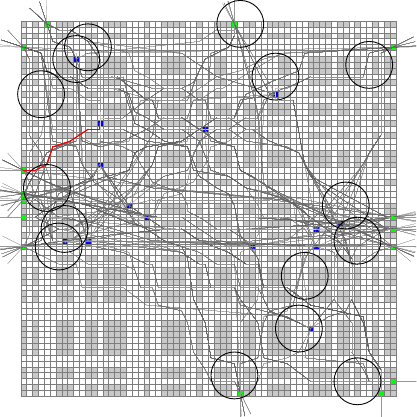}}
\subfloat[\label{fig:newtown:nash:ea}]{\includegraphics[width=.5\textwidth]{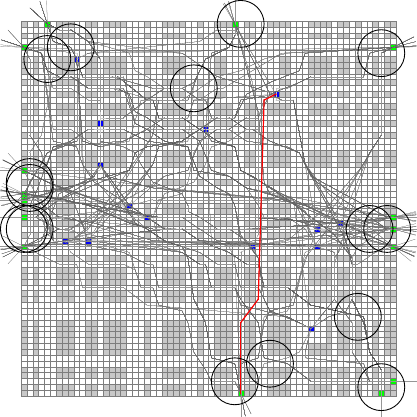}}

\caption{Illustrative solutions found by each of the algorithms on a \newtown\ instance with 15 detectors. (a)
Greedy (b) HC (c) TS (d) EA. Notice the critical path (shown in red) is different in each case. \label{fig:newtown:nash}}
\end{figure*}

As a result of the previous considerations, there is a difference between the relative behavior of
the different iterative algorithms in each of the two scenarios. In the first one (proportional selection),
HC is significantly better than the other algorithms (Quade test $p$-value $\approx 0$, Holm test results
shown in \refTable{tab:holm_prop}). However, in the worst-case equilibrium scenario the EA provides
better results (Quade test $p$-value $\approx 0$, Holm test results
shown in \refTable{tab:holm_nash}). Indeed, if we perform a head-to-head comparison between 
HC and EA in this latter scenario by means of a signrank test, we observe that the EA is significantly better with $p$-value = 0.0057.
Quite interestingly, if we break down the analysis by map class, we find that HC is negligibly and 
non-significantly better than EA in the \harbour\ class ($p$-value = 0.7422), and that the EA is 
significantly better than HC in \newtown\ and \oldtown\ ($p$-values of 0.0098 and 0.0020 respectively).
As a matter of fact, the quantitative difference between the algorithms is clear if we observe the distribution
of deviations from the best known solutions (\refFig{fig:boxplot}): the difference in favor
of HC is much smaller in the proportional selection scenario when compared to the difference in
favor of EA in the worst-case equilibrium scenario. Clearly, the more complex nature of the search
landscape is better dealt with via the diversification of a population-based approach than the strong
intensification of the HC algorithm considered.

\refFig{fig:newtown:nash} 
shows an example of the solutions provided by either algorithm on
a \newtown\ map for the worst-case equilibrium scenario. The solutions provided by each algorithm
are structurally different. The greedy approach tends to place the detectors in the vicinity of the objectives.
The local search approaches seem to have found solutions in which detectors are usually placed close to
entrances, but with some detectors being placed next to objectives as well. Finally the EA has also favored
locations close to entrances in order to cover a large number of paths; however, it has also identified a few
crossroads that seem to be important in order to maximize coverage of many paths. Notice how for
each of these solutions, the resulting critical path is different in each case.

\section{Conclusions and Future Work}
\label{sec:conclusions}

Counterterrorism is a multidimensional endeavor that spreads along many entangled fronts. One of them is 
undoubtedly the tactical defense against targeted attacks, from which suicide bombings constitute a heinous 
distinguished example. While low-rank individuals engaged in international terrorism groups may be often fanaticized and
hence can have motivations that challenge rational analysis (or motivations that, while rational, can be directed
towards other personal or social goals than those of the terror group as a whole \citep{abrahms08terrorists}),
 their actions are often dictated by or inspired by the 
guidelines emanating from the upper levels of a command hierarchy, more prone to engage in analytical
considerations of cost-effectiveness in pursue of their criminal goals. As a matter of fact, the assessment of
the situation would not qualitatively change if rather than considering a suicide attacker embarked on a
no-return mission we consider the case of an unmanned attack using a drone or any other suitable 
vehicle \citep{patterson10unmanned}, a scenario slanted towards purely tactical considerations. In any case, 
assuming this operational context provides at the very least a worst-case scenario upon which a better informed, 
more down-to-earth strategy could be built with the help of 
adequate intelligence. The availability of cogent tools to aid in the analysis of these scenarios can thus be 
most helpful. In this sense, we have considered in this work the use of different heuristic approaches to 
determine the most appropriate deployment of sensors so as to detect perpetrators of bombing attacks in
a given threat area, aiming to minimize the number of casualties or economic damages. Modeling this
situation as an adversarial game in which the attacker is cognizant of the location of these detectors and
tries to target the most cost-effective objective turns out to pose a challenging optimization task. Among the
algorithms considered and tested on a variety of problem scenarios, an evolutionary algorithm turns out
to be quite effective. It is however remarkable that a simple hill climber performs reasonably well on some
instances, suggesting that an eventual hybridization of both methods into a memetic approach \citep{Neri2012book}
might be very promising.

Future work will be actually directed in the direction of hybridizing methods, aiming to exploit their synergy.
Pushing the limits of the resulting techniques (in terms of, e.g., the size of the maps, the number of objectives, 
etc.) would be of paramount interest. Also, it is conceivable to consider in the context described in this work 
extensions of the underlying model including different types of detectors, cf. \citep{Yan201671}, or even
introducing a dynamic component in the configuration of the ground map.



\end{document}